\documentclass{article}
\usepackage[preprint]{colm2026_conference}

\usepackage{microtype}
\usepackage{xcolor}
\usepackage{hyperref}
\usepackage{url}
\usepackage{booktabs}
\usepackage{graphicx}
\usepackage{multirow}

\definecolor{darkblue}{rgb}{0, 0, 0.5}
\hypersetup{colorlinks=true, citecolor=darkblue, linkcolor=darkblue, urlcolor=darkblue}

\title{The Replay Gap: Static Evaluation of Model\\Switching in LLM Agents Scores the Wrong World}

\author{Ashritha Gonuguntla \\
Carnegie Mellon University \\
\texttt{agonugun@andrew.cmu.edu}}

\begin{document}
\maketitle

\begin{center}
\vspace{-1.5em}
\small Accepted at the Efficient Reasoning Workshop, COLM 2026.\\
Code: \url{https://github.com/AshrithaG/replay-gap} \quad
Data: \url{https://huggingface.co/datasets/ashritha0907/replay-gap-trajectories}
\end{center}

\begin{abstract}
LLM routers promise efficiency by matching each request to the cheapest adequate model, and are increasingly applied \emph{per step} inside multi-step agents. Yet agentic routers are evaluated like single-turn routers: by replaying logged trajectories and substituting another model's recorded outputs, assuming the rest of the trajectory is unaffected. We test this assumption with \emph{branching rollouts}: we fork live SWE-bench agent trajectories at controlled points, rebuild the environment, continue each fork with a different model, and compare against same-model control forks that isolate sampling and replay noise. Across six paired runs ($\sim$900 rollouts), swaps exceed their matched control floors by $+0.25$ to $+0.66$ normalized edit distance (multiplicity-corrected CIs exclude zero), rewriting 61--94\% of post-fork actions; 74--77\% of early swaps diverge at the \emph{first} post-fork action, versus 6--35\% of controls --- leaving only 3\% of replayed states valid. Divergence decreases with fork depth in both directions. All five outcome flips we observe occur in swap arms --- upgrades rescuing unsolved instances, a downgrade losing the sole solve --- and zero occur across 359 control forks. Scoring these same swaps with a log-stitching replay evaluator, replay mispredicts \emph{every} success-relevant outcome call and predicts patches with 0.00--0.11 similarity to reality. Auditing the noise floor, temperature-0 ``determinism'' is configuration-dependent: FP8-served controls diverge on over 90\% of forks while AWQ-served ones remain near-identical; and under tight budgets the \emph{stronger} model more often exhausts its steps without submitting. Replay-based benchmarks score the wrong world for agentic routing; we release our harness and all trajectories.
\end{abstract}

\section{Introduction}

Model routing has moved from research prototype to infrastructure. Routers that select among heterogeneous LLMs per query power commercial products and open frameworks \citep{ong2024routellm,chen2023frugalgpt,hu2024routerbench}, and a rapidly growing line of work pushes the routing decision \emph{inside} multi-step agents: selecting a model per step of a software-engineering, tool-use, or computer-use trajectory \citep{vllm2026saar,switchcraft2026,cua2026steplevel}, or jointly selecting models and reasoning strategies under budgets \citep{pan2025rtr}.

Nearly all of this work is evaluated on \emph{logged} model outputs. A benchmark collects each candidate model's answer to each query once; a router is then scored by looking up the answers it would have chosen \citep{hu2024routerbench,huang2025routereval,routerarena2025}. For single-turn queries this is sound: the query does not depend on the router's choice. For agents it is not. An agent's action at step $k$ determines the observation at step $k{+}1$; the trajectory is a closed loop through the environment. Replay evaluation silently assumes an open loop --- that if a router had substituted model $B$ at step $k$, the remainder of model $A$'s logged trajectory would still have happened.

How wrong is this assumption? That trajectories diverge \emph{at all} is not surprising --- an agent is a closed loop, and divergence is the null hypothesis. What routing research needs, and what no one has measured, is the \emph{magnitude and structure} of the divergence: how fast, how far above the noise floor of the serving stack itself, how it depends on swap direction and position, and whether it reaches outcomes. Measuring this requires \emph{branching live rollouts} --- actually forking the trajectory at the decision point and rolling each candidate forward in its own environment --- which is far more expensive than replay. This paper pays that cost and reports what replay hides.

\paragraph{Contributions.}
\begin{enumerate}
\item \textbf{Methodology.} A branching-rollout protocol (and open harness) for counterfactual evaluation of per-step model switching: fork a live trajectory at step $k$, re-execute the action prefix in a fresh container (with logged replay fidelity), seed the message history, and continue with a different model --- always paired with a \emph{same-model control fork} that isolates sampling and environment-replay noise (\S\ref{sec:method}).
\item \textbf{Quantifying the replay gap.} On SWE-bench Verified \citep{jimenez2024swebench}, across two swap directions, two fork positions, and three difficulty/prompt tiers ($\sim$900 rollouts): swaps rewrite 61--94\% of post-fork actions, exceeding matched controls by $+0.25$ to $+0.66$ paired edit distance (95\% CIs exclude zero; Table~\ref{tab:divergence}, Appendix~\ref{app:perrun}); 74--77\% of early swaps diverge at the first post-fork action, leaving 3--8\% of replayed states valid (\S\ref{sec:results}). All five observed outcome flips occur in swap arms; zero in 359 control forks. Scoring the same fixed switch policies with a log-stitching replay evaluator --- exactly the seed-matched logged data such an evaluator would use --- replay mispredicts every success-relevant outcome and its predicted patches are near-orthogonal to the patches actually produced (\S\ref{sec:stitch}).
\item \textbf{Structure with routing implications.} Divergence decreases with fork depth in both directions; upgrades diverge immediately while late downgrades approach control behavior; and we observe the stronger model exhausting tight step budgets without submitting more often than the weak one --- a possible \emph{thoroughness tax} that budget-aware routers may need to price (\S\ref{sec:structure}).
\item \textbf{An audit of ``deterministic'' evaluation.} At temperature 0, control forks of our AWQ-served 14B remain near-identical (edit distance 0.16--0.23; half never diverge), while our FP8-served 4B's controls diverge on 90--96\% of forks (edit distance 0.49--0.67) under identical settings --- so even same-model replay is unsound on some serving stacks. We also document how naive patch-level metrics are inflated by empty-vs-empty comparisons, and report subset-qualified numbers (\S\ref{sec:audit}).
\end{enumerate}

\section{Related Work}

\paragraph{Routing and its benchmarks.} Cost-quality routing spans cascades \citep{chen2023frugalgpt,aggarwal2024automix}, learned per-query routers \citep{ong2024routellm,ding2024hybrid}, RL-trained routers \citep{li2025routerr1}, and routing to unseen models \citep{jitkrittum2025uniroute}. Evaluation infrastructure --- RouterBench \citep{hu2024routerbench}, RouterEval \citep{huang2025routereval}, RouterArena \citep{routerarena2025}, RouteJudge \citep{routejudge2026} --- scores routers against precollected outcomes. Our results do not question these designs for single-turn routing; we show they do not transfer to agents.

\paragraph{Agentic and per-step routing.} Session- and step-level routing has been deployed in serving stacks \citep{vllm2026saar} and studied for tool-calling and computer-use agents \citep{switchcraft2026,cua2026steplevel}; TwinRouterBench \citep{twinrouterbench2026} benchmarks agentic routers with static and ``live dynamic'' evaluation, where \emph{dynamic} refers to changing model availability and prices --- the counterfactual-trajectory problem we study is explicitly left open. Joint selection of models and reasoning budgets \citep{pan2025rtr,odar2026} inherits the same evaluation gap once applied per step.

\paragraph{Off-policy evaluation.} Reinforcement learning formalizes exactly the failure mode we measure: a \emph{behavior} policy generates the logged trajectories, a \emph{target} policy (here, the router's switch) induces a different state distribution, and naive replay is the degenerate estimator that ignores the shift entirely. The standard corrections carry directly: per-decision importance sampling \citep{precup2000eligibility} reweights logged returns by the target/behavior action-probability ratio; weighted and doubly robust variants \citep{jiang2016doubly,thomas2016dataefficient} trade bias for variance by combining reweighting with a learned value model. Applying them to agentic routing raises two obstacles our data speaks to. First, horizon: over 50-step trajectories the product of per-step ratios collapses the effective sample size, precisely the regime where importance sampling is known to degenerate --- and our measured divergence (first divergence at post-fork action ${<}1$ for early upgrades) says the ratios are far from 1 immediately. Second, support: an LLM policy's action space is unbounded text, so the behavior policy assigns vanishing probability to most target actions, violating the coverage assumption these estimators need. This makes branched ground truth valuable beyond measurement --- it is the only way to \emph{validate} whichever estimator one adopts. We report the gap here and leave estimator design to future work.

\paragraph{Nondeterminism in LLM serving.} Batching- and kernel-induced nondeterminism at temperature 0 has been reported informally in the serving community; our same-model control forks give it an experimental floor in the agentic setting, and show it differs sharply across quantization stacks (\S\ref{sec:audit}).

\section{Method}\label{sec:method}

\paragraph{Setup.} We use the mini-SWE-agent scaffold \citep{minisweagent} --- a minimal bash-only ReAct loop --- on SWE-bench Verified \citep{jimenez2024swebench} with the official per-instance Docker images, a 50-step budget, and a 28k-token context. Our pool contains a small model $S$ (Qwen3-4B-Instruct, FP8) and a large model $L$ (Qwen3-14B, AWQ, thinking disabled) \citep{qwen3}, served by vLLM \citep{kwon2023vllm} on a single 24\,GB GPU at temperature 0. This deliberately mirrors the resource-constrained serving regime where routing matters most.

\paragraph{Branching protocol.} For each instance we run a \emph{base} trajectory with the base model to termination. We then select fork steps at 30\% and 70\% of the base trajectory's length. For each fork: (i) start a fresh container; (ii) re-execute the recorded actions of all pre-fork steps, logging return-code agreement with the recorded observations (\emph{replay fidelity}: across 11{,}702 replayed actions in 708 branches, 99.99\% of return codes match and 707/708 branches reconstruct exactly --- Appendix~\ref{app:fidelity}, so environment reconstruction error is not a plausible source of the divergence we report); (iii) seed the agent's message history with the recorded prefix, so the branch model sees exactly what the base model saw; (iv) continue the rollout with the branch model to termination. Each fork point receives two arms: a \textbf{swap arm} (the other model) and a \textbf{same-model control arm}, which absorbs sampler nondeterminism, batching effects, and environment-replay drift. Divergence attributable to the model swap is read \emph{relative to the control}.

\paragraph{Runs.} A \emph{run pair} is one 30-instance sweep in one swap direction: forward (base $S$, swap up to $L$) or reverse (base $L$, swap down to $S$). We execute six run pairs --- \{full difficulty, easy bucket (``$<$15\,min fix''), easy with a budget-nudged prompt\} $\times$ \{forward, reverse\}, seed-matched --- yielding 717 scored branch pairs from $\sim$900 rollouts.

\paragraph{Metrics.} Post-fork action edit distance (Levenshtein over exact command strings, normalized by the \emph{longer} of the two post-fork suffixes; since suffix lengths differ systematically by arm, we complement it with two length-insensitive statistics: the first divergent action index and the fraction of branches diverging at all); \emph{replay validity}, the prefix-match fraction (share of the base's post-fork actions before first divergence --- the states a replay evaluator would score correctly); patch metrics --- file-set Jaccard, \emph{patch similarity} (character-level \texttt{SequenceMatcher} ratio between the raw unified-diff texts, in $[0,1]$), and exact identity --- all subset-qualified (\S\ref{sec:audit}); official SWE-bench resolution per arm; and exit-status distributions.

\section{Results}\label{sec:results}

\begin{figure}[t]
\centering
\includegraphics[width=\linewidth]{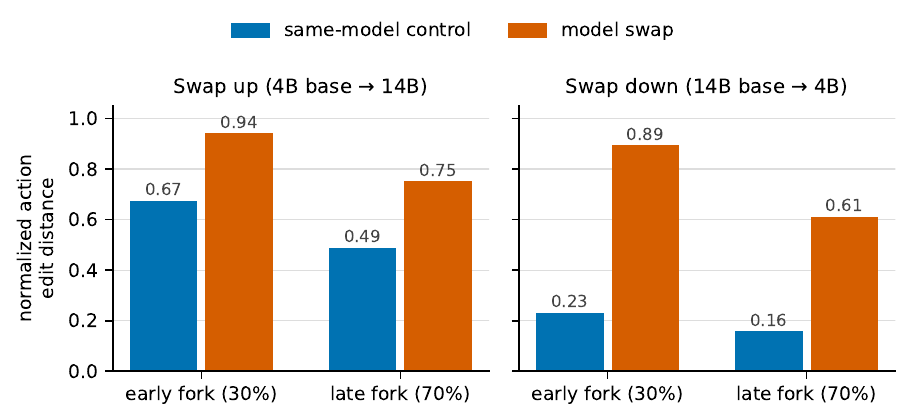}
\caption{Post-fork action divergence by direction and fork position, pooled over three run pairs per direction. Swaps (orange) diverge above same-model controls (blue) in every arm; divergence decreases with fork depth; upgrades diverge more than downgrades. Per-run-pair values in Appendix~\ref{app:perrun}.}
\label{fig:divergence}
\end{figure}

\begin{table}[t]
\centering
\small
\begin{tabular}{llrrrrr}
\toprule
direction & arm & $n$ & edit dist. & \% diverged & 1st div. & replay validity \\
\midrule
\multirow{4}{*}{up ($S{\to}L$)}
 & control@early & 88 & 0.674 & 95.5\% & 4.9 & 27.3\% \\
 & control@late  & 91 & 0.489 & 90.1\% & 6.3 & 62.7\% \\
 & \textbf{swap@early} & 88 & \textbf{0.941} & \textbf{100.0\%} & \textbf{0.6} & \textbf{3.2\%} \\
 & swap@late     & 90 & 0.752 & 98.9\% & 3.6 & 34.4\% \\
\midrule
\multirow{4}{*}{down ($L{\to}S$)}
 & control@early & 90 & 0.232 & 53.3\% & 17.4 & 73.8\% \\
 & control@late  & 90 & 0.158 & 50.0\% & 11.0 & 85.8\% \\
 & \textbf{swap@early} & 90 & \textbf{0.895} & \textbf{97.8\%} & \textbf{2.1} & \textbf{8.0\%} \\
 & swap@late     & 90 & 0.611 & 83.3\% & 4.1 & 39.4\% \\
\bottomrule
\end{tabular}
\caption{Pooled post-fork divergence (three run pairs per direction). ``1st div.''\ is the mean index of the first divergent action among diverged branches; \emph{replay validity} is the fraction of post-fork states a replay evaluator would score against the correct world. Paired swap$-$control deltas: up@early $+0.267$ [$0.190, 0.349$], up@late $+0.254$ [$0.189, 0.317$], down@early $+0.663$ [$0.580, 0.744$], down@late $+0.453$ [$0.374, 0.534$] (bootstrap 95\% CIs; resampling is at instance level --- each instance contributes exactly one control$-$swap delta per arm, so branches within an instance are never split across resamples). All four deltas also exclude zero under Bonferroni correction for the four arms (98.75\% CIs: lower bounds $+0.172$, $+0.175$, $+0.560$, $+0.347$).}
\label{tab:divergence}
\end{table}

\subsection{The replay gap at the action level}

\begin{figure}[t]
\centering
\begin{minipage}{0.48\linewidth}
\includegraphics[width=\linewidth]{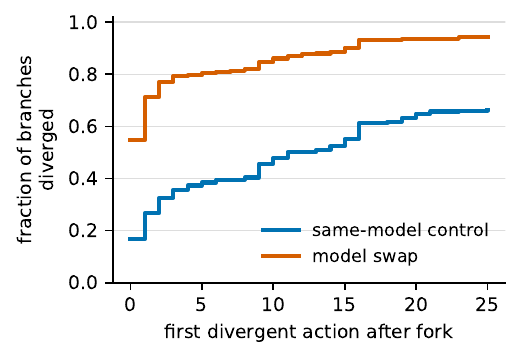}
\end{minipage}\hfill
\begin{minipage}{0.48\linewidth}
\includegraphics[width=\linewidth]{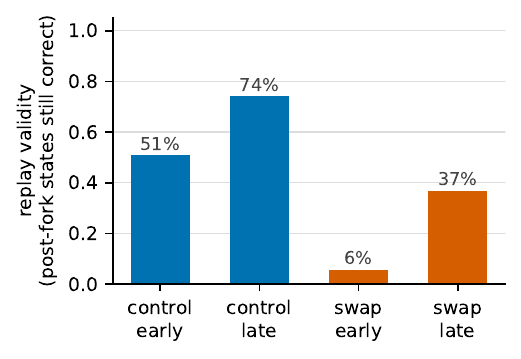}
\end{minipage}
\caption{\textbf{Left:} ECDF of the first divergent post-fork action (pooled over arms): 55\% of all swap branches --- and 74--77\% of early swaps --- diverge at action 0. \textbf{Right:} replay validity by arm --- the share of post-fork states replay evaluation scores correctly.}
\label{fig:validity}
\end{figure}

Table~\ref{tab:divergence} and Figure~\ref{fig:divergence} give the core result. In every direction and position, swap arms diverge above their matched controls, and the paired per-instance deltas are significant in all four arms (95\% bootstrap CIs exclude zero; table caption). Because the forward direction's FP8-served base is itself noisy (\S\ref{sec:audit}), its control floor is high and the ceiling-bounded delta ($+0.27$) understates the effect; \textbf{the reverse direction, whose AWQ-served controls are near-deterministic, is our cleanest evidence: swapping down at the early fork adds $+0.66$ edit distance over a $0.23$ floor.} Early swaps diverge at the \emph{first} post-fork action in 73.9\% (up) and 76.7\% (down) of branches --- against control action-0 rates of 35.2\% and 5.6\% respectively, an excess of $+38.7$ and $+71.1$ points, so the immediate re-decision is a swap effect, not a noise-floor artifact. The replay-validity column makes the consequence concrete: \textbf{a replay evaluator of an early swap scores 92--97\% of post-fork decisions against a state that never occurs} (Figure~\ref{fig:validity}). The pattern replicates in each of the six run pairs individually (Appendix~\ref{app:perrun}).

\subsection{Outcome flips are exclusive to swap arms}

Absolute resolution rates are low in our constrained regime (base rates 0--3\%; \S\ref{sec:limitations}), so we report outcome flips exactly as observed: \textbf{five flip events across three instances, all in swap arms, and zero flips across 359 same-model control branches.} At full difficulty, an early upgrade resolved an instance the small model never solved in any arm (the late upgrade of the same instance did not). In the nudged easy tier, upgrades resolved \texttt{django\_\_django-11163} at \emph{both} fork positions. In the easy reverse tier, swapping down lost the one instance the large model had solved, at both fork positions, while its control retained it at both. Thus two of the three flip instances were fork-position invariant. Replay evaluation is structurally unable to observe any of these events, because none of them exist on the logged trajectory.

\subsection{Structure: position, direction, and the thoroughness tax}\label{sec:structure}

Three regularities matter for routing policy. \textbf{(i) Position:} divergence decreases with fork depth in both directions (two positions tested) --- late handoffs inherit the base's completed work through the workspace, making late downgrades far safer than early ones. \textbf{(ii) Direction:} upgrades diverge immediately (the stronger model re-decides at once), downgrades diverge later and less. \textbf{(iii) Budgets (an observation, not yet a regularity):} in the full-difficulty run pair, the \emph{large} model exhausted the 50-step budget without submitting more often than the small one (24/30 vs 17/30 base rollouts; two-proportion $z\!\approx\!2.0$, uncorrected, single run pair) --- consistent with it exploring and verifying more per task, though ``this budget starves this model on this stack'' is an equally available reading. If it generalizes, routing \emph{up} under tight budgets can reduce completion rate --- a thoroughness tax worth validating in competent regimes. A prompt-level budget reminder in the easy tier shifted submissions in opposite directions for the two models (8\,$\to$\,11 and 10\,$\to$\,6 of 30; counting noise at this $n$), suggesting budget control itself consumes the context budget.

\subsection{Auditing the noise floor and patch metrics}\label{sec:audit}

Two audit findings qualify common evaluation practice. \textbf{Temperature-0 determinism is configuration-dependent:} control forks of the AWQ-quantized 14B remain near-identical to their base trajectories (edit distance 0.16--0.23; half never diverge at any action), while the FP8-quantized 4B's controls diverge on 90--96\% of forks (edit distance 0.49--0.67) under identical decoding settings --- batched FP8 serving is not a deterministic oracle, so even same-model replay carries irreducible noise on some stacks. \textbf{Patch-identity metrics inflate silently:} when neither side submits a non-empty patch, naive patch identity scores 1.0; restricted to branch pairs where both sides submitted real patches, identity drops (e.g., 0.60--0.67 for late arms and 0.00 for early arms in the forward runs, on small $n$). We report all patch-level numbers subset-qualified and recommend the practice.

\subsection{Replay-stitch predictions vs.\ branched ground truth}\label{sec:stitch}

The preceding sections show replay's \emph{inputs} are invalid; here we test whether its \emph{predictions} fail, using no new rollouts. Because our forward and reverse tiers are seed-matched, for every instance we hold both models' full logged standalone trajectories \emph{and} the live branched result of each fixed switch policy (up/down $\times$ early/late). A logged-outcome evaluator in the style of single-turn router benchmarks predicts a switch's result from the target model's logged run on the same instance; we compare those predictions to the branched truth (Table~\ref{tab:stitch}). We do not claim any specific published benchmark implements precisely this stitching rule; we test the \emph{assumption they share} --- that a model's logged performance on an instance transfers to the switched-to position --- which any log-based evaluator of mid-trajectory switching must make in some form. Our result bounds what that assumption can deliver in this regime.

\begin{table}[t]
\centering
\small
\begin{tabular}{llrrrrr}
\toprule
policy & fork & $n$ & outcome agree & missed succ. & false succ. & patch sim. \\
\midrule
swap-up   & early & 88 & 96.6\% & 2/2 & 1 & 0.108 \\
swap-up   & late  & 90 & 97.8\% & 1/1 & 1 & 0.020 \\
swap-down & early & 89 & 100.0\% & 0/0 & 0 & 0.033 \\
swap-down & late  & 89 & 100.0\% & 0/0 & 0 & 0.000 \\
\bottomrule
\end{tabular}
\caption{A log-stitching replay evaluator vs.\ branched ground truth. ``Outcome agree'' is inflated by the majority class (nearly all rollouts fail); the decisive calls are the success-relevant ones. ``Patch sim.''\ is the mean similarity between the replay-predicted patch and the patch the switch actually produced, over pairs where either is non-empty. $n$ differs slightly from Table~\ref{tab:divergence} because a branch is scoreable here only if its instance's seed-matched standalone run exists (one easy-tier instance has none, removing one branch per down arm).}
\label{tab:stitch}
\end{table}

Raw outcome agreement is 97--100\% --- and vacuous, since predicting universal failure is almost always right in this regime. The decisive calls tell the real story: \textbf{replay predicted failure for all three switching successes that actually occurred, and both successes it did predict never materialized --- 0-for-5 on every prediction where success was at stake.} These five decisive calls are \emph{not} the five flip events of \S4.2: the two downgrade losses are cases replay happens to call correctly (both models' logged runs fail, and so does the switch), while the two false successes are prediction errors invisible at the flip level; the three missed rescues are common to both counts. At the patch level, replay's predicted patch has similarity 0.00--0.11 to the patch the switch actually produces. For calibration, even a model predicting \emph{itself} from logs (its base patch vs.\ its own same-model control branch) reaches only 0.31 (FP8 4B) to 0.53 (AWQ 14B) --- so any log-based evaluator is capped by serving-stack nondeterminism before the cross-model error is even added. (These ceilings are computed over the same either-patch-non-empty subset as the rest of the column; the never-diverged empty-vs-empty control pairs that make the AWQ controls near-identical in \S\ref{sec:audit} are excluded, which is why 0.53 coexists with that determinism.) For a null reference: the trivial always-predict-failure evaluator scores 2/5 on these same calls (it is right wherever replay hallucinated a success) --- so the logged information made the stitch evaluator \emph{strictly worse than a constant predictor}. The success count is small (five decisive events) and we claim no more than the sign of the result; but the sign is uniform, and it is the direction replay's structural blindness predicts.

\section{Implications}\label{sec:implications}

\textbf{For evaluation:} agentic router benchmarks need live or branched evaluation; static replay scores 92--97\% of post-fork decisions against invalid states for early swaps (61--97\% across all swap arms), cannot observe outcome flips at all, and --- tested directly on our own policies --- mispredicted every success-relevant outcome (\S\ref{sec:stitch}). Our harness and branched dataset provide a starting point, and off-policy estimators over logged trajectories (importance sampling, doubly robust) are the natural cheap approximation to validate against it. \textbf{For routers} (stated as hypotheses our regime motivates but cannot confirm --- \S\ref{sec:limitations}): fork position and direction look like first-class features --- hand off downward late, once the work is done; escalate hard instances early, before the trajectory ossifies; and price a model's step appetite, not only its per-step quality. Validating these in competent regimes is the immediate next step.

\section{Limitations}\label{sec:limitations}

This is a deliberately controlled pilot: one scaffold, one benchmark family, one model family (two quantized members), $n{=}30$ instances per run pair at temperature 0, and a 24\,GB serving budget (28k context) that keeps absolute resolution rates low (0--3\%). Action-level divergence and the one-sided flip evidence do not require competent agents to be valid --- replayed states are off-distribution regardless of task success --- but we show the \emph{inputs} to replay evaluation are invalid rather than demonstrating end-to-end router mis-ranking; measuring mis-ranking in competent regimes, with unquantized pools and a third fork position, is the immediate next step alongside OPE-corrected estimators. Quantization differences (FP8 vs AWQ) entangle model capability with serving stack --- our up/down comparisons are therefore between \emph{deployment configurations}, not pure model scales. Prefix replay assumes command-level environment determinism, which we verify rather than assume (99.99\% return-code agreement over 11{,}702 replayed actions; Appendix~\ref{app:fidelity}), and the branch model inherits the base model's in-context style through the replayed prefix, which is part of the phenomenon rather than an artifact, but merits study.

\section*{Reproducibility}
The branching harness, all run configurations, and every analysis script behind the numbers in this paper are released at \url{https://github.com/AshrithaG/replay-gap}, and the full $\sim$900-rollout branched-trajectory dataset (base and fork trajectories, per-step actions and observations, fork metadata, replay-fidelity logs, token counts, patches, and SWE-bench outcomes) at \url{https://huggingface.co/datasets/ashritha0907/replay-gap-trajectories}.

\bibliography{refs}
\bibliographystyle{colm2026_conference}

\appendix

\section{Per-run-pair divergence}\label{app:perrun}

Mean post-fork normalized action edit distance per run pair. The swap-above-control ordering holds in all 24 cells.

\begin{table}[h]
\centering
\small
\begin{tabular}{llrrrr}
\toprule
direction & run pair & ctrl@early & ctrl@late & swap@early & swap@late \\
\midrule
\multirow{3}{*}{up ($S{\to}L$)}
 & full difficulty & 0.634 & 0.468 & 0.932 & 0.789 \\
 & easy            & 0.669 & 0.481 & 0.951 & 0.759 \\
 & easy, nudged    & 0.717 & 0.518 & 0.940 & 0.706 \\
\midrule
\multirow{3}{*}{down ($L{\to}S$)}
 & full difficulty & 0.192 & 0.093 & 0.910 & 0.606 \\
 & easy            & 0.233 & 0.168 & 0.859 & 0.653 \\
 & easy, nudged    & 0.272 & 0.214 & 0.917 & 0.575 \\
\bottomrule
\end{tabular}
\caption{Replication across tiers: the swap arm exceeds its matched control in every cell of every run pair.}
\label{tab:perrun}
\end{table}

\section{Prefix-replay fidelity}\label{app:fidelity}

Every branch re-executes the base trajectory's pre-fork actions in a fresh
container; we compare each replayed action's return code against the one
recorded in the base trajectory. Table~\ref{tab:fidelity} reports the
agreement. Environment reconstruction is near-exact --- one mismatched action
in 11{,}702, and 707 of 708 branches replayed perfectly --- so the post-fork
divergence reported in \S\ref{sec:results} cannot be attributed to replay
error in the prefix.

\begin{table}[h]
\centering
\small
\begin{tabular}{lrrrr}
\toprule
run pair & branches & replayed actions & return-code match & branches exact \\
\midrule
full difficulty, up   & 116 & 1{,}524 & 100.00\% & 100.0\% \\
full difficulty, down & 120 & 2{,}624 & 100.00\% & 100.0\% \\
easy, up              & 116 & 1{,}500 & 100.00\% & 100.0\% \\
easy, down            & 116 & 2{,}112 & 99.95\%  & 99.1\% \\
easy nudged, up       & 120 & 1{,}660 & 100.00\% & 100.0\% \\
easy nudged, down     & 120 & 2{,}282 & 100.00\% & 100.0\% \\
\midrule
\textbf{all}          & \textbf{708} & \textbf{11{,}702} & \textbf{99.99\%} & \textbf{99.9\%} \\
\bottomrule
\end{tabular}
\caption{Prefix-replay fidelity. ``Branches exact'' is the fraction of branches in which every replayed action's return code matched the recording.}
\label{tab:fidelity}
\end{table}

\end{document}